\documentclass[letterpaper, 10 pt, conference]{ieeeconf}  
\usepackage{tikz}

\IEEEoverridecommandlockouts                              

\usepackage{graphicx}      
\usepackage{caption}
\usepackage{subcaption}
\usepackage{amsmath}
\usepackage{amssymb}
\usepackage{xcolor}
\usepackage{import}

\newtheorem{lemma}{Lemma}

\usepackage{multirow}
\usepackage{booktabs}

\usepackage{algorithm2e}
\usepackage{mathtools, cases}
\usepackage{cuted} 
\usepackage{placeins}
\usepackage{bm}

\usepackage[left=1.91cm,right=1.91cm,top=1.91cm,bottom=1.91cm]{geometry}

\makeatletter
\let\NAT@parse\undefined
\makeatother
\usepackage[hidelinks]{hyperref} 

\title{\LARGE \bf
A Corrector-aided Look-ahead Distance-based Guidance for Online Reference Path Following with an Efficient Mid-course Guidance Strategy}

\author{Reva Dhillon$^{1*}$, Agni Ravi Deepa$^{1*}$, Hrishav Das$^{2^{*,+}}$, Subham Basak$^1$, Satadal Ghosh$^3$
\thanks{$^{*}$These authors contributed equally to this work.}
\thanks{$^{1}$Students, Department of Aerospace Engineering, Indian Institute of Technology Madras.
        \tt\small revasdhillon@gmail.com, jrdps4123@gmail.com, ae20d412@smail.iitm.ac.in }
\thanks{$^{2}$Student, Department of Aerospace Engineering, University of Illinois Urbana–Champaign. 
        \tt\small hrishav.das@gmail.com}
\thanks{$^{+}$Majority of the work has been done while Hrishav was at IIT Madras.}
\thanks{$^{3}$Associate Professor, Department of Aerospace Engineering, Indian Institute of Technology Madras.
        \tt\small satadal@iitm.ac.in}}

\begin{document}

\maketitle

\begin{abstract}
Efficient path-following is crucial in most of the applications of autonomous vehicles (UxV). Among various guidance strategies presented in literature, the look-ahead distance ($L_1$)-based nonlinear guidance has received significant attention due to its ease in implementation and ability to maintain a low cross-track error while following simpler reference paths and generating bounded lateral acceleration commands. However, the constant value of $L_1$ becomes problematic when the UxV is far away from the reference path and also produces higher cross-track error while following complex reference paths having high variation in radius of curvature. To address these challenges, the notion of look-ahead distance is leveraged in a novel way to develop a two-phase guidance strategy. 
Initially, when the UxV is far from the reference path, an optimized $L_1$ selection strategy is developed to guide the UxV towards the vicinity of the start point of the reference path, while maintaining minimal lateral acceleration command. Once the vehicle reaches a close neighborhood of the reference path, a novel notion of corrector point is incorporated in the constant $L_1$-based guidance scheme to generate the guidance command that effectively reduces the root mean square of the cross-track error and lateral acceleration requirement thereafter. Simulation results validate satisfactory performance of this proposed corrector point and look-ahead point pair-based guidance strategy, along with the developed mid-course guidance scheme. Also, its superiority over the conventional constant $L_1$ guidance scheme is established by simulation studies over different initial condition scenarios.






\end{abstract}

\section{INTRODUCTION}


With advancements in technology, autonomous vehicles are gaining widespread adoption in both defense and civilian sectors. One key component in these applications is path following. For instance, unmanned vehicles (UxVs) are required to follow a reference path having sensing and positional advantage in ISR missions, while self-driving cars are required to follow / change lanes autonomously on the go. Thus, development of efficient guidance algorithms for precise path-following is essential. 

In existing literature, extensive research has been carried out for devising guidance strategies for different types of reference path following. Traditional proportional-integral-derivative (PID) controller-based methods \cite{rhee} demonstrate effective performance when dealing with small heading and cross-track errors within the linear domain. However, their effectiveness reduces while following complicated reference paths. On the other hand, nonlinear approaches, such as 
sliding mode-based \cite{labbadi2021adaptive},\cite{nian20232}, 
nonlinear model predictive control \cite{faulwasser2015nonlinear},
virtual target-based backstepping technique \cite{lapierre2006nonsingular},
methods are more adept at following complex trajectories.
However, these guidance strategies are highly dependent on the system model and tend to be complex in implementation \cite{doi:10.2514/1.G001060}. 
To  the contrary, vector field-based methods \cite{nelson2007} have emerged as a promising approach to address this limitation by using cross-track-error-dependent smooth field structure to define desired course angles and/or speeds for smooth convergence toward a reference path. 
However, most of the existing vector-field-based methods are designed for specific curve types, such as straight lines or circular arcs, making them unsuitable for general curves. To this end, a vector field-based method was presented in \cite{basak2025general} to suit it for generic 3-D reference trajectories. Nevertheless, these methods do not inherently ensure bounded guidance commands and can generate guidance formulation that is complex to implement.

In contrast, guidance-theoretic path-following techniques are straightforward, independent of specific models, and easy to implement \cite{doi:10.2514/1.G001060}. 
These include Pure Pursuit (PP) \cite{morales2009pure}, Line-of-Sight Guidance (LOSG) \cite{rysdyk2006unmanned}, Proportional Navigation (PN) \cite{dhananjay2012guidance} and hybrid approaches combining PP and LOSG \cite{kothari2014uav}. A PN-inspired nonlinear guidance strategy was introduced in \cite{park2007} to guide a vehicle by following a virtual point located at a fixed look-ahead distance $L_1$ along the desired path. A three-dimensional nonlinear guidance law was presented in \cite{doi:10.2514/1.G001060}, using differential geometry of 3-D space curves. Among these nonlinear techniques, the look-ahead-based guidance method gained prominence due to its simplicity, bounded lateral acceleration command, and asymptotic stability. Several studies have further refined this approach. In \cite{PARK201264}, the acceleration command given by the outer-loop is modified with the gravity term being subtracted for better control in the inner-loop. Another 3-D guidance method is discussed in \cite{9610125}, wherein the 3-D problem is decomposed to generate the guidance command, along the longitudinal and lateral planes, and the constant $L_1$ guidance scheme of \cite{park2007} is applied to both the planes.

Although a fixed look-ahead distance $L_1$ ensures that the guidance command remains bounded, this constant value of $L_1$ can lead to higher cross-track errors when following paths with varying curvature radii. To address this, a variable $L_1$ scheme was presented in \cite{SG_Variable_L1_icuas}, where the look-ahead distance is adjusted based on a pre-defined allowable settling time and maximum cross-track error.
Besides, a neural network-based adaptive $L_1$ guidance law was introduced in \cite{8899601} for enhanced performance in path following. 

However, dynamically adjusting the $L_1$ value causes frequent variations in lateral acceleration commands, which affect actuator performance. Besides, a small value of $L_1$ can lead to high lateral acceleration commands, which may be impractical for real-world implementation. Moreover, these guidance strategies, being specifically meant for the homing phase, are not directly effective when the UxV is far from the reference path. This also necessitates the development of an effective mid-course phase guidance. To this end, this paper first introduces an effective mid-course phase to guide the UxV to a sufficiently close vicinity of the planar reference path in an optimal way. Then, another guidance command is developed, in which a fixed look-ahead distance aided by a novel notion of corrector point is leveraged, which facilitates a reduction in cross-track error during close-range complex curved path following in 2-D. Considering their advantages of look-ahead distance-based guidance formulation is also adopted for both the phases in this paper. Numerical simulations are presented to validate the effectiveness of the developed path following guidance strategy in varied initial condition scenarios. 

The paper is organized as follows: Section~\ref{sec:PbmDefn} states the problem description and some prerequisites for subsequent sections. Section~\ref{sec:Meth} details the presented guidance strategies. Section~\ref{sec:Sims} shows the simulation results, followed by the concluding remarks in section~\ref{sec:Conclusion}.

\section{PROBLEM DEFINITION AND BACKGROUND}
\label{sec:PbmDefn}
\subsection{Problem Definition \label{sec:Problem Definition}}
For a 2-D motion of an unmanned vehicle (UxV) modeled as a point mass moving at a constant speed $V$, with heading angle $\psi$ relative to the $x$-axis of the inertial frame, the kinematic equations are given as:
\begin{equation}
\label{kinematiceqs}
\begin{aligned}
\dot{x}=V \cos \psi;\:\:\dot{y}=V \sin \psi;\:\:\dot{\psi}={a_v}/{V}
\end{aligned}
\end{equation}
Here, $a_v$ is the lateral acceleration input to the UxV. The inner loop controller is assumed ideal implying that the commanded lateral acceleration ($a_{cmd}$) is always equal $a_v$. Therefore, $\dot{\psi}={a_{cmd}}/{V}$. 
All angles and angular velocities are considered positive in the anti-clockwise sense. 

\begin{figure}[]
    \centering
    \includegraphics[width=0.8\linewidth]{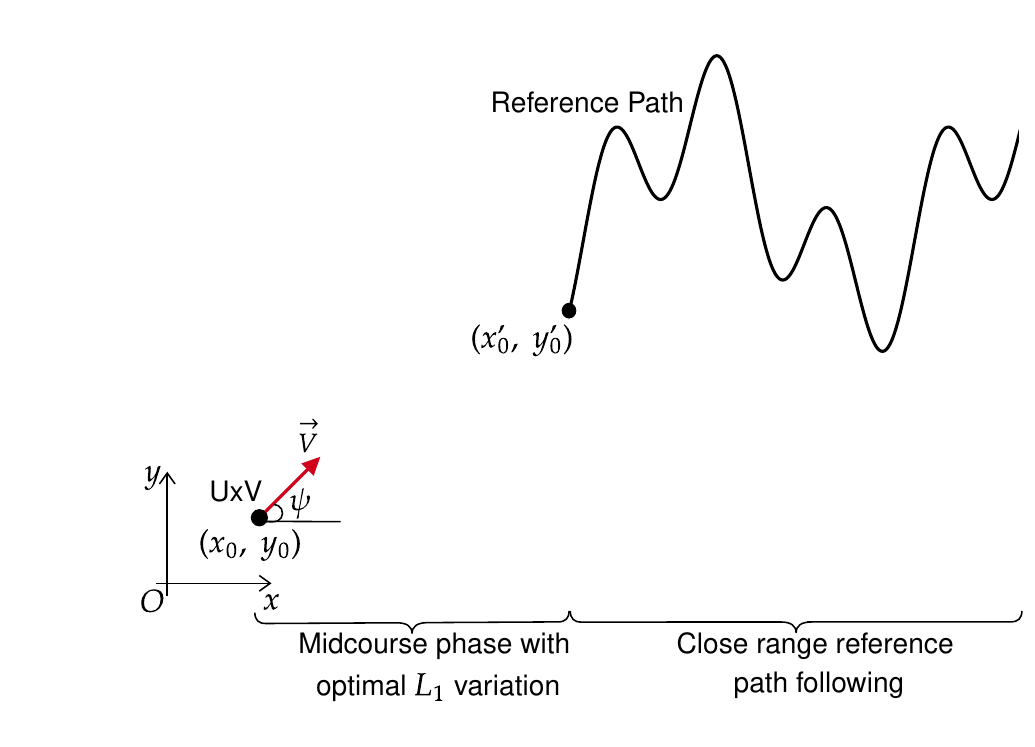}
    \caption{Initial engagement geometry}
    \label{fig:Pbm_defn}
\end{figure}

In this paper, a {$C^2$ continuous reference path} is considered, with the assumption that the entire reference path is pre-defined. Given this, a guidance algorithm is to be developed , which would generate suitable lateral acceleration command $a_{cmd}$ to ensure that the UxV with kinematics represented in \eqref{kinematiceqs} follows the desired path as closely as possible, starting from a far-away location, as shown in Fig. \ref{fig:Pbm_defn}.



\subsection{Background Results }
\label{subsec_background}
The constant $L_1$ path following guidance scheme presented in \cite{park2007} considers a look-ahead point that is always at a distance of $L_1$ from the UxV on the reference path. As this point moves along the path, the UxV gradually steers toward it, thereby finally closely tracking the reference path.
The guidance command is defined as follows,
\begin{equation}
    a_{cmd} = 2\frac{V^2}{L_1}\sin{\eta}.
    \label{acceleration command}
\end{equation}
Here, $L_1$ is the distance between the UxV's current position and the look-ahead point (virtual target) and $\eta$ is the angle between its velocity vector and the line of sight (LOS) vector between the virtual target and the UxV.
Note that the guidance command given in \eqref{acceleration command} directs the UxV along a circular arc joining the UxV position and the look-ahead point such that the instantaneous UxV velocity is tangent to the circular arc. This command can accurately track a circle with $L_1\leq d$, where $d$ is the diameter of the circular arc. 

\begin{figure}[!h]
    \centering
    \includegraphics[width=0.6\linewidth]{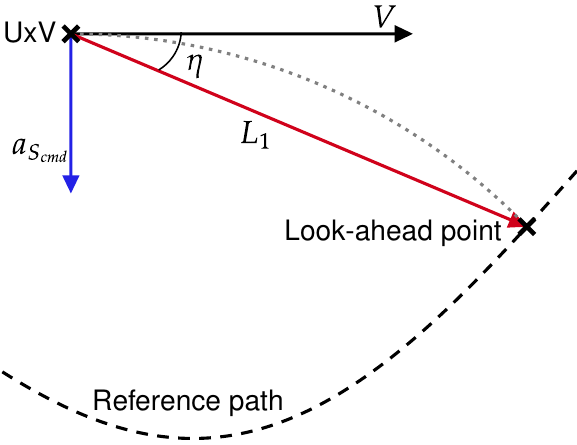}
    \caption{Engagement Geometry for constant Look-Ahead Guidance ($\eta<0$ illustrated)}
    \label{fig:Diagram explaining the acceleration command}
\end{figure}

The method received significant attention because of its ease of implementation and anticipative nature, which enables tight-tracking of curved paths. It incorporates the ground speed of the UxV, which leads to good performance even in the presence of disturbances. It is proven to be a stable guidance scheme for small deviations from the reference path. The nonlinear guidance method is also shown to be asymptotically Lyapunov stable for circular paths (Refer to Section IV of \cite{park2007}).


\section{DEVELOPMENT OF THE CORRECTOR-AIDED PATH FOLLOWING GUIDANCE ALONG WITH MID-COURSE PHASE STRATEGY}\label{developmentofguidance}

As mentioned in Section \ref{sec:Problem Definition}, the UxV's initial location $(x_0,y_0)$ is considered to be far away from $(x_0',y_0')$, the starting point of the reference path, that is $\sqrt{(x'_0 - x_0)^2 + (y'_0-y_0)^2} \geq 2 R_0$. One of the shortcomings of the constant $L_1$ guidance discussed in Section~\ref{subsec_background} is its inability to guide the UxV when it is located at more than twice the radius of curvature at the initial point on the reference path $R_0$. Hence, it is essential to devise guidance strategies for both the mid-course phase and terminal phase for the given reference path following. For that, the notion of look-ahead distance-based guidance is first utilized in a novel way to direct the UxV to the start of the reference path in an optimal manner during the mid-course phase, as described in Section \ref{midcourse_section}.  

Upon reaching a close vicinity of the start of the reference path, the close-range path following phase commences. In the close-range path following phase, the accuracy of the constant $L_1$ guidance method decreases as the path shifts from circular arcs to more complex curves. It can be observed that to traverse a circular path, $\Dot{\psi}$ must be constant and $\Ddot{\psi} = 0$. Whenever $\Ddot{\psi} \neq 0$ is required, as in the case of general curved trajectories, a higher cross-track error is incurred by the constant $L_1$ strategy. To this end, a novel notion of utilizing a corrector point along with the constant distance look-ahead point to improve the accuracy of path following is presented in Section \ref{Sec:Terminal Path Following Guidance}. 




\label{sec:Meth}
\subsection{Mid-course Guidance with Optimal Selection of $L_1$}
\label{midcourse_section}
For the mid-course phase, an intermediate circle, hereafter referred to as 'initiation circle', with nominal kino-dynamically feasible radius of turn $R$ is considered such that it touches the reference path tangentially at its start point. The look-ahead distance-based guidance philosophy is leveraged in this section to reach this initiation circle tangentially from a far away initial position. Upon reaching the initiation circle, the UxV follows it to reach the start point of the reference path along the desired direction. 


\subsubsection{Dynamic Look-Ahead Distance for Optimally Moving to the Initiation Circle}
\label{movingtocircle}



\begin{figure}[!h]
    \centering
    \includegraphics[width=0.7\linewidth]{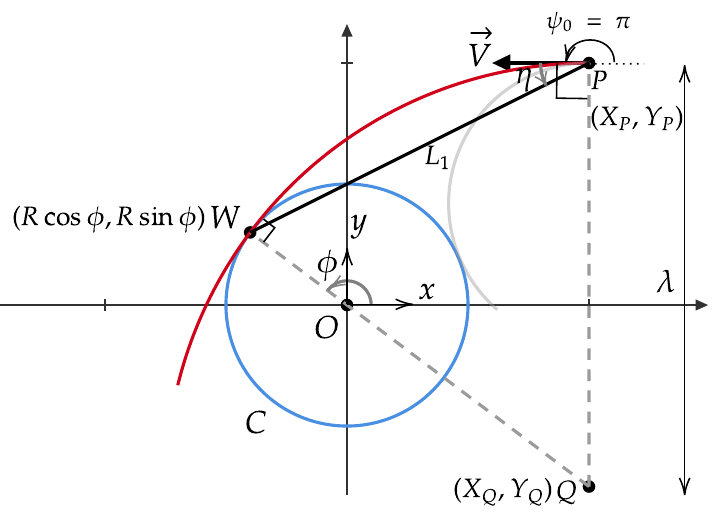}
    \caption{Moving to Initiation Circle}
    \label{fig:Phase 1 Explanation schematic}
\end{figure}

Consider a scenario in Fig. \ref{fig:Phase 1 Explanation schematic}, where the UxV starts from point $P$ with initial heading $\psi_0 = \pi$ and needs to be guided to the initiation circle $C$, which is centered at $O$ and of radius $R$. The set of all candidate look-ahead points $(R \cos(\phi), R \sin(\phi))$ lie on $C$, where $\phi\in [-\pi,\pi)$. Given the axi-symmetric nature of circles, this scenario represents all initial points and heading angles without loss of generality.

\lemma
\label{lemma1}
The look-ahead point on the initiation circle $C$ that corresponds to the extremum lateral acceleration (following Eq. \eqref{acceleration command}) for steering the UxV (located at $P$ with velocity $\vec{V}$) to $C$ tangentially is the contact point, at which the circle $C$ touches another circle that passes through $P$ tangentially along $\vec{V}$.


\begin{proof}
Differentiating Eq. \eqref{acceleration command} with respect to (w.r.t.) $\phi$ and enforcing condition for extrema,

\begin{equation}
\label{A_cmd_Derivative}
\begin{aligned}
&\frac{d a_{cmd}}{d\phi} = 2V^2 \left(\frac{\cos(\eta)}{L_1}\frac{d \eta}{d\phi} - \frac{\sin(\eta)}{L_{1}^{2}} \frac{d L_1}{d\phi}\right)=0 \\
\Rightarrow &{(\frac{d \eta}{d\phi})}/{(\frac{d L_1}{d\phi})} = {\tan(\eta)}/{L_1}\\
\end{aligned}
\end{equation} 


It is now to be proved that the point $W$, the tangent contact point as shown in Fig. \ref{fig:Phase 1 Explanation schematic}, satisfies Eq. \eqref{A_cmd_Derivative}. From analyzing the geometry of the scenario and taking derivatives, the following equations are obtained.

\begin{subequations} \label{Phase1Proof1geometry}
\begin{align}
    \tan\phi &= \frac{Y_P-\lambda}{X_P}  \label{phistuff}\\
    \cos\phi &= \frac{X_P}{R-\lambda} \label{cosphistuff}\\
    \lambda &= \frac{R^2 - X_P^2 - Y_P^2}{2 (R-Y_P)} \label{lambdastuff} 
\end{align}
\end{subequations}
where, $\lambda$ is the radius of the bigger circle that passes through $P$ tangentially along $\vec{V}$. The general expressions of $\eta$ and $L_1$ are obtained as:
\begin{subequations}
\label{Phase1Proof1etaL1}
\begin{align}
    \eta &=  \tan^{-1}\left(\frac{R\sin\phi - Y_P}{R\cos\phi - X_P}\right) \label{etastuff}\\
    L_1 &= \sqrt{(R\cos\phi - X_P)^2 + (R\sin\phi - Y_P)^2} \label{L1stuff}
\end{align}
\end{subequations}
Differentiating these expressions w.r.t. $\phi$:
\begin{subequations}
    \label{Phase1Proof1Derivatives}
    \begin{align}
        \frac{d \eta}{d\phi} &= \frac{R^2 - RX_P\cos\phi - RY_P\sin\phi}{L_1^2} \label{detastuff} \\
    \frac{d L_1}{d\phi} &= \frac{R(X_P\sin\phi - Y_P\cos\phi)}{L_1} \label{dL1stuff}
    \end{align}
\end{subequations}
When Eq.~\eqref{Phase1Proof1geometry} is enforced, the conditions in Eq.~\eqref{Phase1Proof1Derivatives} necessarily satisfy the extrema condition of $a_{cmd}$ given in Eq.~\eqref{A_cmd_Derivative}. It can be observed that there are two circles and two potential look-ahead points that satisfy the extrema condition defined. The point $W$ represents the circle with the higher radius of curvature. Hence, this initiation circle corresponds to a minima in terms of acceleration.
\end{proof}

Thus, lemma \ref{lemma1} provides an optimal selection of the reaching point on the initiation circle for the UxV. Based on the UxV's initial position and heading, the initial look-ahead distance is found and used in its guidance command \eqref{acceleration command}, which steers it through the bigger circle. Along this circular trajectory of the UxV during the mid-course phase, the optimal reaching point on the initiation circle remains invariant. Hence, the look-ahead distance $L_1$ varies as the UxV position varies, but the ratio ${\sin(\eta)}/{L_1}$ remains constant.



\begin{figure}[]
    \centering
    \includegraphics[width=0.8\linewidth]{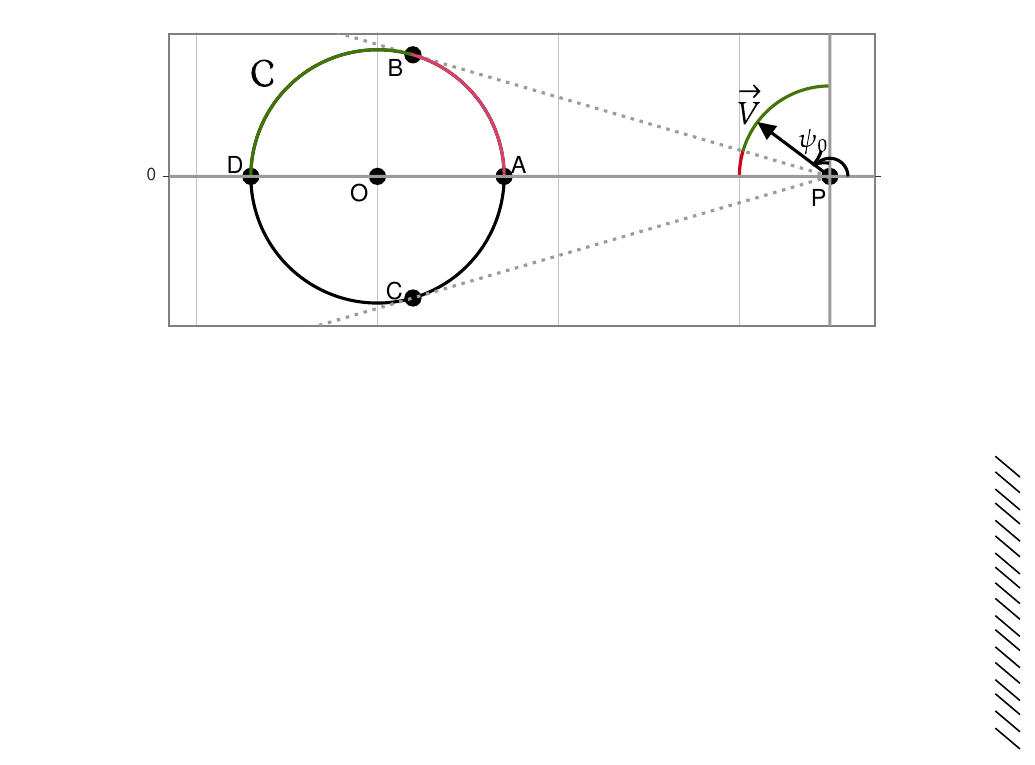}
    \caption{Initial Heading to Contact Point Mapping}
    \label{InferencesfromPhase1Method}
\end{figure}

A few key inferences on how different initial UxV heading angles ($\psi_0$) map to different optimal reaching points ($W$ in lemma \ref{lemma1}) can be drawn from the above-mentioned optimal method of approaching the initiation circle. Referring to Fig. \ref{InferencesfromPhase1Method}, note the followings.

\begin{itemize}
    \item This work considers initial UxV headings wherein the angle between the velocity vector and the line joining the UxV to the circle centre is $\leq 90^\circ$.
    \item If $\psi_0$ directs the UxV in the red sector, the optimal reaching point lies on the red circular arc $AB$.
    \item If $\psi_0$ directs the UxV in the green sector, the optimal reaching point lies on the green circular arc $BD$.
    \item Due to symmetry between second and third quadrants of the UxV-fixed reference frame in terms of heading towards the circle $C$, a similar trend is observed for $\psi_0$ belonging to third quadrant as well.
\end{itemize}



\subsubsection{Initiation Circle Selection} \label{circleselection}
\begin{figure}[]
        \centering
        \begin{subfigure}{0.35\textwidth}
            \centering
            \includegraphics[width=\textwidth]{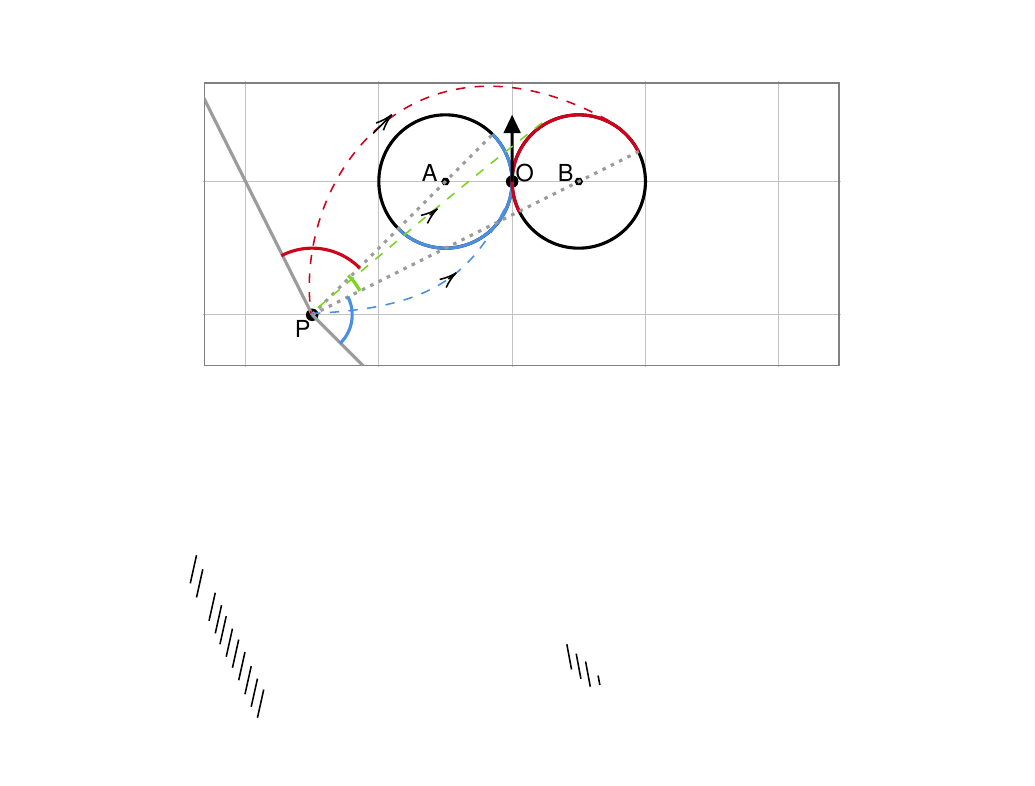} 
            \caption{Case 1}
            \label{fig:PositoveDotScenario}
        \end{subfigure}
        \begin{subfigure}{0.35\textwidth}
            \centering
            \includegraphics[width=\textwidth]{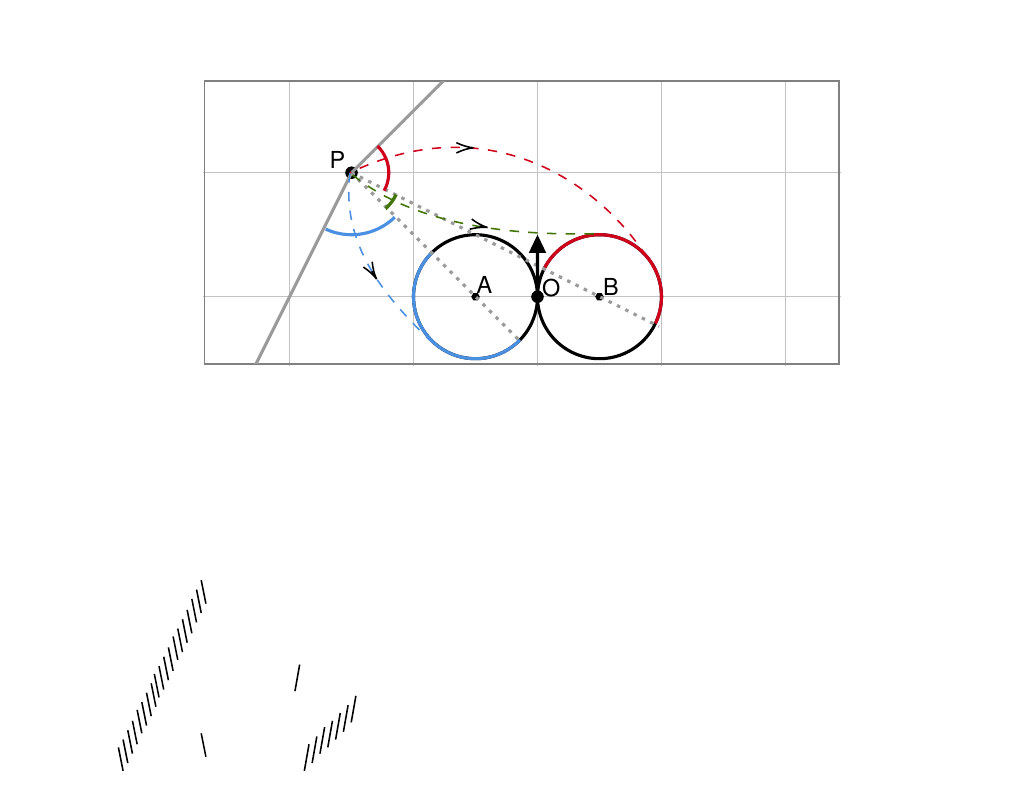} 
            \caption{Case 2}
            \label{fig:NegativeDotScenario}
        \end{subfigure}
        \caption{Circle Selection in Midcourse Phase}
        \label{fig:circle_selection}
    \end{figure}

Given the nominal radius of the initiation circle and the start point of the reference path along with the desired heading direction of the UxV at that point, two candidate initiation circles exist as can be seen in Fig.~\ref{fig:circle_selection}, where $O$ is the start point on the reference path, and the initial heading of the reference path to be tracked is ${\pi}/{2}$ (shown by the black arrow). Any general engagement scenario can be considered to be one of the two cases as shown in Figs. \ref{fig:PositoveDotScenario} and \ref{fig:NegativeDotScenario}. 

\textit{Case 1:} When the initial UxV location $P$ lies behind the start point of the reference path. (Fig. \ref{fig:PositoveDotScenario}).
\begin{itemize}
    \item If $\psi_0$ directs the UxV in the red sector, the optimal reaching point lies on the red half of the initiation circle centered at $B$.
    \item If $\psi_0$ directs the UxV in the blue sector, the optimal reaching point lies on the blue half of the initiation circle centered at $A$.
    \item If $\psi_0$ directs the UxV in the green sector, both the circles provide feasible lowest $a_{cmd}$ option for the UxV to reach the start point of reference path along the desired heading. Among them, the initiation circle is selected which demands a lesser lateral acceleration.
    \item The red, blue and green dotted lines in Fig. \ref{fig:PositoveDotScenario} depict the UxV trajectories based on this selection criteria.
\end{itemize}

\textit{Case 2:} When the initial point $P$ lies in front of the start point of the trajectory. (Fig. \ref{fig:NegativeDotScenario}).
\begin{itemize}
    \item If $\psi_0$ directs the UxV in the red sector, the optimal reaching point lies on the red half of the initiation circle centered at $B$.
    \item If $\psi_0$ directs the UxV in the blue sector, the optimal reaching point lies on the blue half of the initiation circle centered at $A$.
    \item If $\psi_0$ directs the UxV in the green sector, none of the circles gives the desired orientation following lowest $a_{cmd}$ circular arcs. Instead, we revert to \eqref{Phase1Proof1geometry} and consider the alternate circle (shown in light grey in Fig. \ref{fig:Phase 1 Explanation schematic}). Once again, we pick between the two circles based on lower lateral acceleration demands among them.
    \item The red, blue and green dotted lines in Fig. \ref{fig:NegativeDotScenario} depict the UxV trajectories based on this selection criteria.
\end{itemize}


\subsection{Close-Range Reference Path Following Guidance}\label{Sec:Terminal Path Following Guidance}
Once the UxV reaches within close range of the reference path after the mid-course phase following an optimal look-ahead point-based guidance as discussed in Section \ref{midcourse_section}, the close-range path following commences. For this phase, a novel corrector point-aided look-ahead distance method is introduced in this section. Selection criteria for the corrector point based on the UxV location and local reference path information, and the resulting guidance command for close-range reference path following are detailed below.

\subsubsection{Obtaining the Corrector Point} \label{findcorrector} 
\begin{figure}[]
    \centering
    \includegraphics[width=0.8\linewidth]{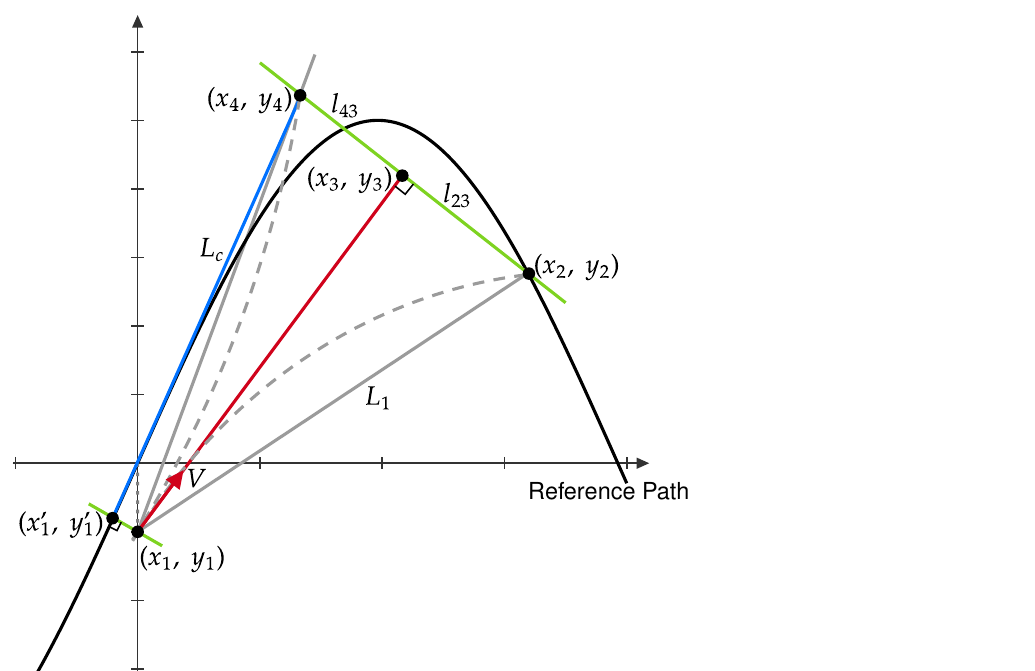}
    \caption{Corrector and Look-ahead Points Pair Schematic.}
    \label{fig:Corrector pseudo-target pair schematic}
\end{figure}

Fig.~\ref{fig:Corrector pseudo-target pair schematic} illustrates the method following which the corrector point is selected. Consider that the UxV is located at $(x_1, y_1)$, which does not lie on the path. First, its projection on the reference path is obtained as $(x'_1, y'_{1})$ (closest point projection). Else, if the UxV lies on the reference path, then $(x'_1, y'_{1}) = (x_1, y_1)$. Next, a line (green) perpendicular to the direction of the velocity $\Vec{V}$ and passing through the look-ahead point $(x_2, y_2)$ is constructed. The look-ahead point is a suitable $L_1$ distance away from $(x_1,y_1)$ on the path. Then, the tangent (blue) to the reference path at $(x'_1, y'_{1})$ is constructed, which intersects the green straight line at $(x_4, y_4)$, which is named as 'corrector point' in this paper. While $L_1$ is the look-ahead distance of the original look-ahead point, $L_c$ is that of the corrector point. From its formulation stated above and from Fig. \ref{fig:Corrector pseudo-target pair schematic}, it is evident that even though the look-ahead point tends to fail closely following the reference path, the corrector point attempts to hold the UxV heading closely following the reference path by embedding the local reference path information in it.  

\subsubsection{The Resulting Guidance Command}\label{subsubsec:acc_new} The mathematical form of the guidance command is given as follows.

\begin{equation}
\label{eq:acc_new}
    a_{cmd} = {(w_1a_{12} + w_2a_{14})}/{(w_1 + w_2)}
\end{equation}
where, 
\begin{equation}
    w_1 = {k_1R_{L1}}/{(1 + l_{23})}
\end{equation}
\begin{equation}
    w_2 = {k_2v_m}/{(R_{L1}(1 + l_{43}))}
\end{equation}
\begin{equation}
\label{acceleration command:12} 
    a_{12} = {2V^2\sin{\eta_{12}}}/{L_{1}}
\end{equation}
\begin{equation}
\label{acceleration command:14} 
    a_{14} = {2V^2\sin{\eta_{14}}}/{L_{c}}
\end{equation}
Here, $l_{ij}$ denotes the distance between two points,
\begin{equation}
    l_{ij} = \sqrt{(x_i - x_j)^2 + (y_i - y_j)^2};\:\:i,j\in\{1,2,3,4\};\:\:i\neq j
\end{equation}
And, the radius of curvature of the path at $(x_2, y_2)$ is given by $R_{L1}$,
\begin{equation}
\label{eq:R}
    R_{L1} = \left. {\left[ 1 + \left(\frac{dy}{dx}\right)^2\right]^{\frac{3}{2}}}/{\frac{d^2y}{dx^2}}  \right|_{(x_2, y_2)}
\end{equation}
and $v_l$ is the speed of the look-ahead point at $(x_2, y_2)$ found by enforcing that $L_1$ is considered constant (analogous to $V_T$ in Section 3A in \cite{park2007}). Also, $v_m\triangleq\frac{V + v_l}{2}$. The angles $\eta_{12}$ and $\eta_{14}$ are measured from the UxV heading direction to the direction of look-ahead and corrector points, respectively. The quantities $k_1$ and $k_2$ are non-negative guidance parameters. They are found by using an optimizer to minimize the cost defined by the root mean square (RMS) of the cross-track error over the reference path locally. This optimization of guidance parameters is detailed later in Section \ref{sec:Online Optimization}.
%
Now, the effects of the weights of the guidance command are described below:
\begin{itemize}
    \item $w_1$: The weight given to the look-ahead point-based guidance command $a_{12}$ decreases if the radius of curvature $R_{L1}$ at $(x_2, y_2)$ is lower. The factor of $(1+l_{23})$ in the denominator ensures that a look-ahead point closer to $(x_3, y_3)$ is weighed more than that at a larger distance, since the lateral acceleration demand would be more for a look-ahead point at a larger distance.
    \item $w_2$: The weight given to the corrector point increases if $R_{L1}$ is lower and $v_l$ is higher, as these two values are indicative of a greater curvature than that of a circle. The factor ${v_m}/{R_{L1}}$ is indicative of the turn rate of the look-ahead point. Hence, if the turn rate of the look-ahead point increases, the weight given to the corrector point increases. The role of $(1+l_{43})$ is the same as that of $(1+l_{23})$ in $w_1$.
\end{itemize}

The lateral acceleration command \eqref{acceleration command:12} from the look-ahead point directs the UxV along the arc joining $(x_1, y_1)$ and $(x_2, y_2)$. For the corrector point, the lateral acceleration command \eqref{acceleration command:14} directs the UxV along the arc joining $(x_1, y_1)$ and $(x_4, y_4)$. Therefore, their weighted average as given in \eqref{eq:acc_new} enables the UxV to generate a guidance command that ensures the UxV follows the desired path more accurately. The entire corrector-aided look-ahead point-based path following guidance scheme is summed in Algorithm \ref{algorithm}.



\subsubsection{Online Optimal Selection of Guidance Parameters $k_1$ and $k_2$}\label{sec:Online Optimization}
The optimization for the weights in \eqref{eq:acc_new} is carried out in an online manner by minimizing a performance index based on locally computed cross-track error over guidance parameters $k_1$ and $k_2$ at every time interval $\Delta t$, where this time interval is computed adaptively in sync with the local path curvature as follows:
\begin{itemize}
    \item Compute radius of curvature $R_{\text{proj}}$ at $(x_1', y_1')$.
    \item Set a distance $d_{online} = \min({d_{limit}, R_{\text{proj}}})$, where the value $d_{limit}$ can be set to the maximum allowed path length, for which a particular set of $k_1$, $k_2$ are permissible. We set $d_{limit}= 2L_1$. 
    \item Compute $\Delta t = d_{online}/V$.
    \item At every Guidance time-interval, the differential equations \eqref{kinematiceqs} are propagated over $\Delta t$ time (computed above), and RMS of cross-track error over this $\Delta t$ time is computed for different values of allowable $k_1$ and $k_2$.
    \item The values of $k_1$ and $k_2$ are selected based on minimization of locally computed RMS cross-track error (as stated above).
    \item This online optimization scheme facilitates optimizing the path following performance by considering the optimization over a short path-segment when the path curvature is high and a large segment in case of low path curvature. 
\end{itemize}

\subsubsection{The Overall Guidance Algorithm}\label{sec:Overall Guidance Algorithm} 
Thus, a novel guidance scheme is presented above to efficiently reach the reference path with optimal acceleration, and closely track it with the coefficients required for tracking being updated in an online manner. It is represented below in the following algorithm.



\RestyleAlgo{ruled}
\begin{algorithm}[!h]
\caption{Corrector-aided Look-ahead Distance-based Guidance} \label{alg:One}
\label{algorithm}
\textbf{Obtain:} Initial UxV position $(x_0,y_0)$, velocity $V$, initial heading angle $\psi_0$ and details of the reference path.\\ 

\While{Not End of Path}{
  \eIf{ $\sqrt{(x'_0 - x_0)^2 + (y'_0-y_0)^2} \geq 2 R_0$ is TRUE}{
    
   Construct two candidate initiation circles (Section \ref{circleselection});\\
   Select correct initiation circle (Section \ref{circleselection});\\
   Find contact point $W$ on this circle (Section \ref{movingtocircle});\\
   Compute $a_{cmd}$ from Eq. \eqref{acceleration command} with $W$ as a static look-ahead point and look-ahead distance $L_{1_{\text{midcourse}}} =  \sqrt{(x'_0 - x_0)^2 + (y'_0-y_0)^2}$. Use this $a_{cmd}$ till the end of midcourse phase;\\
   Update UxV position $(x_1,y_1)$, $\psi$ from Eq.~\eqref{kinematiceqs};

  }{Fix guidance parameters:$L_{1_\text{terminal}}$\\
  \While {$(x_2,y_2) \neq$ \text{End of path}}{
    Find corrector point (Section \ref{findcorrector});\\
    Compute distances $L_c$, $l_{23}$ and $l_{43}$ (Section \ref{subsubsec:acc_new}).\\
    Compute $v_m$ and $R_{L1}$ (Section \ref{subsubsec:acc_new}) ;\\
    Update $a_{cmd}$ from Eq.~\eqref{eq:acc_new};\\
    Update UxV position $(x_1,y_1)$, $\psi$, from Eq.~\eqref{kinematiceqs};\\
    Update look-ahead point position $(x_2,y_2)$;}
  }}
  
\end{algorithm}

\section{SIMULATION RESULTS}
\label{sec:Sims}

\begin{table*}
\centering
\caption{Comparison of Control Methods Across Initial Heading Angles}
\label{tab:method_comparison}
\begin{tabular}{lccc ccc ccc}
\toprule
\textbf{Initial Heading} 
& \multicolumn{3}{c}{\textbf{Fixed L1 Method}} 
& \multicolumn{3}{c}{\textbf{Proposed Method}} 
& \multicolumn{3}{c}{\textbf{Relative Improvement (\%)}} \\
& $a_{\text{cmd, rms}}$ & $d_{\text{rms}}$ & $a_{\text{cmd, max}}$ 
& $a_{\text{cmd, rms}}$ & $d_{\text{rms}}$ & $a_{\text{cmd, max}}$ 
& $a_{\text{cmd, rms}}$ & $d_{\text{rms}}$ & $a_{\text{cmd, max}}$ \\
\midrule
$a_1 = -20.882^\circ$  & 1.715 & 1.211 & 4.830 & 1.495 & 1.190 & 4.830 & 12.838 & 1.734  & 0.000 \\
$a_2 = -5.882^\circ$   & 1.536 & 1.007 & 4.330 & 1.405 & 0.967 & 4.330 & 8.528  & 3.972  & 0.000 \\
$a_3 = 9.119^\circ$    & 1.465 & 0.847 & 3.536 & 1.324 & 0.782 & 3.536 & 9.641  & 7.674  & 0.000 \\
$a_4 = 24.118^\circ$   & 1.402 & 0.718 & 2.500 & 1.256 & 0.643 & 2.500 & 10.423 & 10.321 & 0.000 \\
$a_5 = 39.118^\circ$   & 1.330 & 0.632 & 2.293 & 1.205 & 0.473 & 2.455 & 9.398  & 25.158 & -7.060 \\
$a_6 = 54.118^\circ$   & 1.336 & 0.609 & 2.293 & 1.180 & 0.532 & 2.519 & 11.674 & 12.640 & -9.865 \\
$a_7 = 69.118^\circ$   & 1.326 & 0.654 & 2.294 & 1.180 & 0.575 & 2.458 & 11.018 & 12.080 & -7.134 \\
$a_8 = 84.118^\circ$   & 1.347 & 0.760 & 2.294 & 1.204 & 0.693 & 2.458 & 10.612 & 8.947  & -7.147 \\
$a_9 = 99.118^\circ$   & 1.401 & 0.912 & 2.294 & 1.250 & 0.850 & 2.465 & 10.775 & 6.806  & -7.471 \\
$a_{10} = 114.12^\circ$& 1.470 & 1.099 & 2.294 & 1.311 & 1.063 & 2.474 & 10.816 & 3.279  & -7.857 \\
$a_{11} = 129.12^\circ$& 1.548 & 1.311 & 2.294 & 1.388 & 1.295 & 2.577 & 10.335 & 1.220  & -12.349 \\
\bottomrule
\end{tabular}
\end{table*}

In this section, the developed guidance strategy presented in Section \ref{developmentofguidance} is validated using numerical simulations performed in MATLAB (R2025a) environment on an AMD Ryzen $9$ $5900$ HS processor. The proposed method is simulated for one sinusoidal path. First, a comparison of its performance is given w.r.t. the baseline constant $L_1$ guidance \cite{park2007} for a fixed initial UxV position, but eleven different initial UxV heading angles. Subsequently, the complete engagement scenario, including both the midcourse phase and close-range path following, is demonstrated through simulation. 


For the mid-course phase, first the contact point $W$ is obtained, and once the UxV completes the mid-course phase as described in Section \ref{circleselection}, it switches to close-range path following as given in Section \ref{Sec:Terminal Path Following Guidance}. For close-range tracking, the constants $k_1, k_2$ are iteratively determined in the online manner at every Guidance time-interval as described therein. The interior point optimizer fmincon (MATLAB) is used for this optimization with maximum iterations set to $1000$ and maximum number of function evaluations set to $10^7$. 

The simulation details are listed below.
\begin{itemize}
\item Desired path: $y=10\sin{(0.078x)}+20\cos{(0.082x)}$ 
\item The constant UxV speed $V = 5\ m/s$ 
\item The constant distance $L_1 = 10\ m$. 
\item These values of $V$ and $L_1$ correspond to a maximum turn rate $\dot{\psi}= 57.30 ^\circ/s$, which is kin-dynamically feasible for a UAV.
\item Initial UxV location $= (-15, 0)$.

\end{itemize}

\subsection{Close-Range Reference Path Following}  

The measures of effectiveness are defined as follow,
\begin{equation}
\begin{aligned}
    CTE_\text{RMS} &= \left(1 - \frac{\text{Cost}(\text{optimized } k_1, k_2)}{\text{Cost}(k_1 = 1, k_2 = 0)}\right) \times 100 \%\\
    AE_\text{RMS} &= \left(1 - \frac{\text{latax-rms}(\text{optimized } k_1, k_2)}{\text{latax-rms}(k_1 = 1, k_2 = 0)}\right) \times 100 \%   
\end{aligned} 
\end{equation} where $CTE_\text{RMS}$ and $AE_\text{RMS}$ denote the improvement in RMS of cross-track error and guidance (lateral acceleration a.k.a. latax) command over the entire reference path. The constant $L_1$ method \cite{park2007} can be recovered from the new proposed form of the acceleration command by setting $k_1 = 1, k_2 = 0$ 

\begin{figure}[]
    \centering
        \includegraphics[width=1\linewidth]{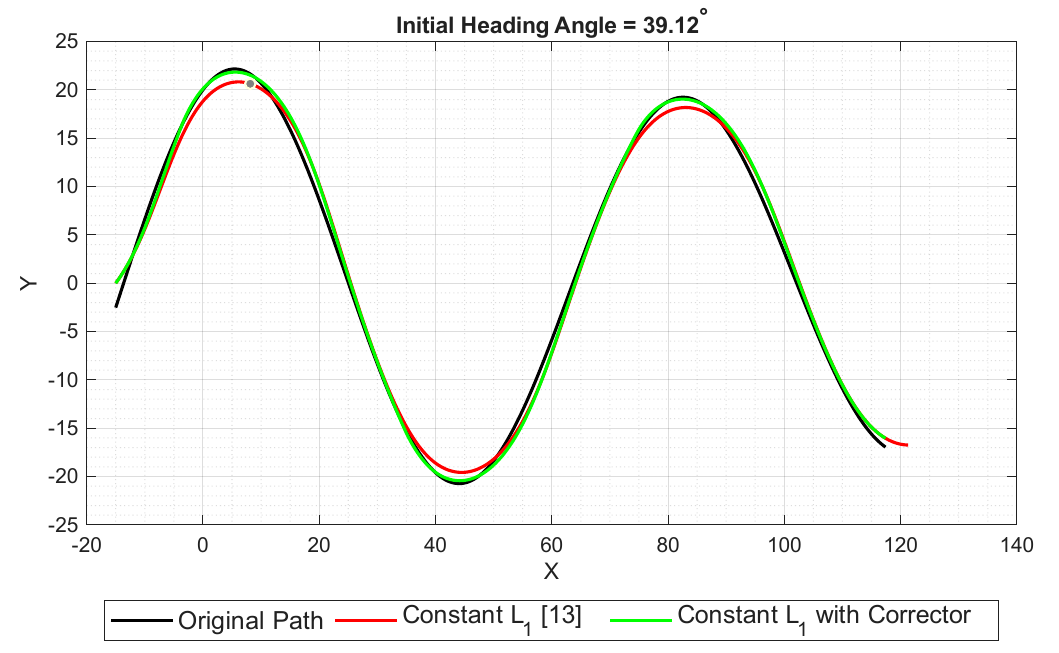}
        \caption{A. Path obtained from the schemes}
        \label{fig:case1}
\end{figure}
\begin{figure}[!h]
    \centering
        \includegraphics[width=1\linewidth]{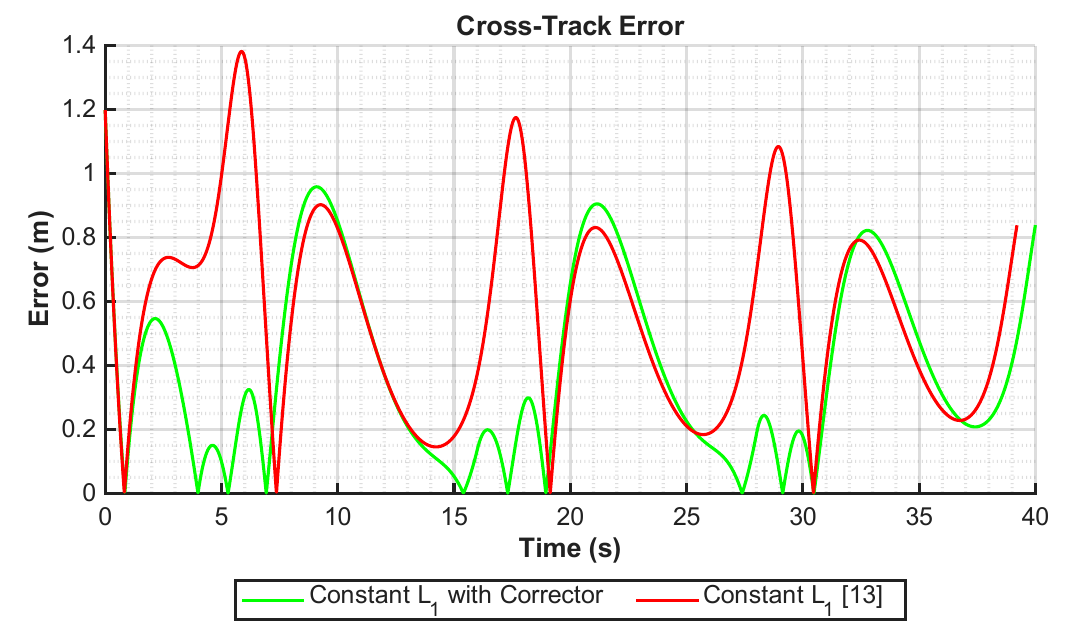}
        \caption{A. Cross-track error variation}
        \label{fig:cross1}
    \end{figure}
\begin{figure}[!h]
    \centering
        \includegraphics[width=1\linewidth]{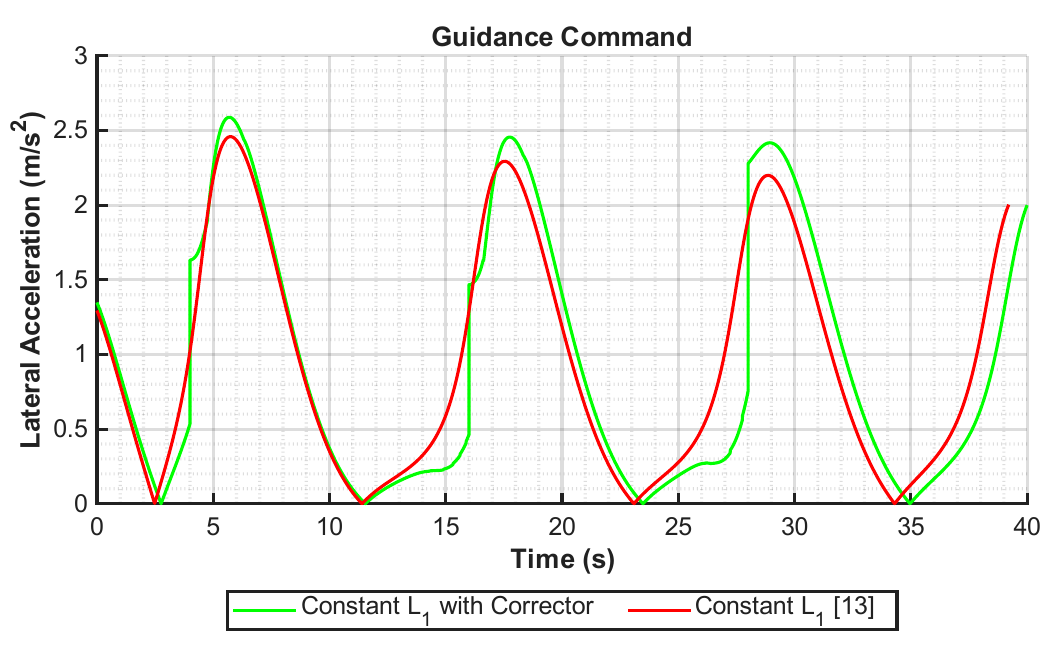}
        \caption{A. Lateral acceleration variation}
        \label{fig:acc1}
    \end{figure}

For initial heading $\psi_0 = 39.118^\circ$, The actual UxV trajectory is plotted in Fig. \ref{fig:case1}, the cross-track error variation is plotted in Fig. \ref{fig:cross1} and the lateral acceleration requirements are plotted in Fig. \ref{fig:acc1}. It can be seen that the corrector-aided constant look-ahead distance-based guidance method presented in this paper allows the UxV to follow the path much more closely. 

Note from Table \ref{tab:method_comparison} that for each simulation, the corrector point aided method outperforms constant $L_1$ guidance for the same value of the look-ahead distance in terms of both of these measures of effectiveness (listed in the last column of the table). 
Evidently, the presented method performs veritably better than the original based on these metrics. 


\subsection{Overall Guidance and Illustration}
\begin{figure}[!h]
    \centering
        \includegraphics[width=1\linewidth]{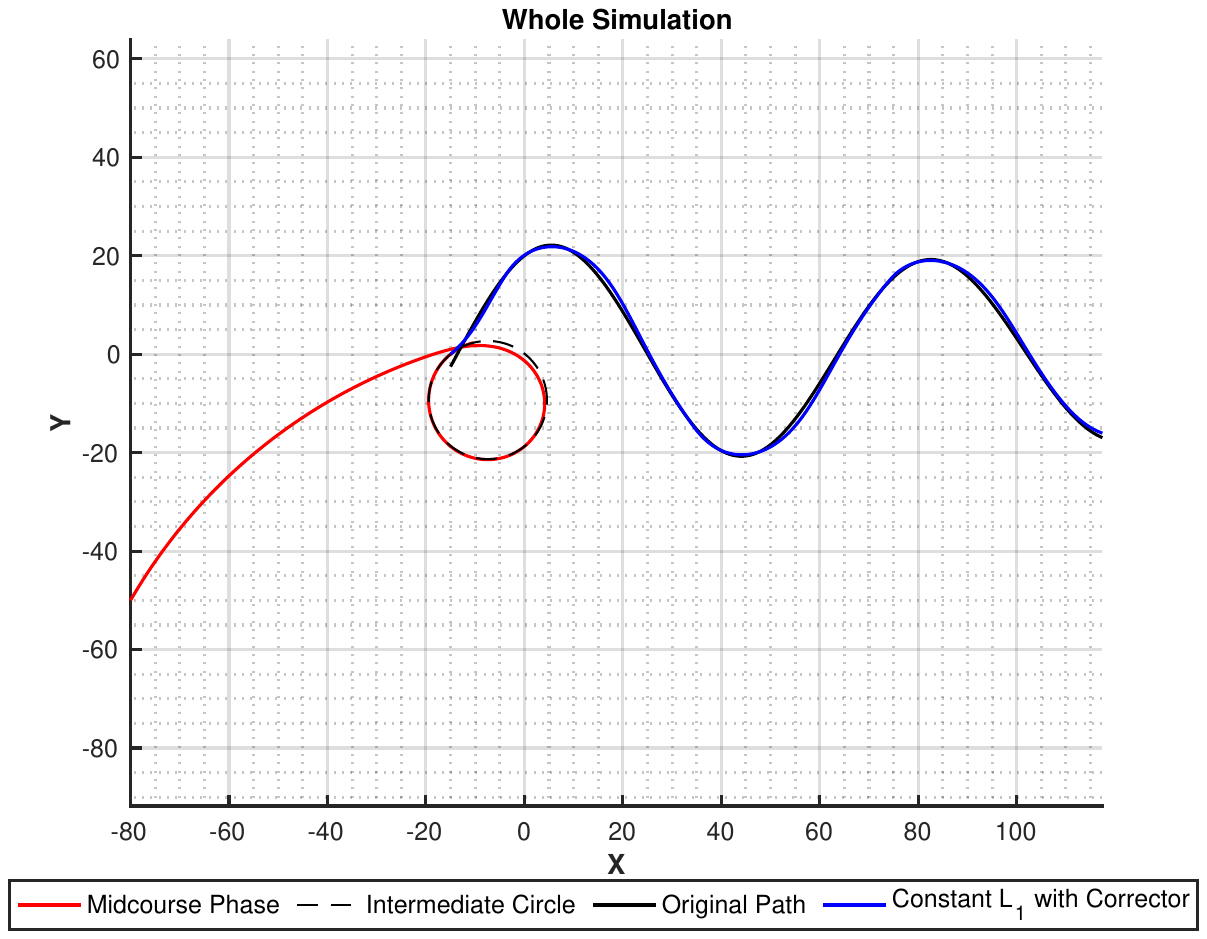}
        \caption{B. Path obtained from the schemes}
        \label{fig:wholesim}
    \end{figure}

\begin{figure}[!h]
    \centering
        \includegraphics[width=1\linewidth]{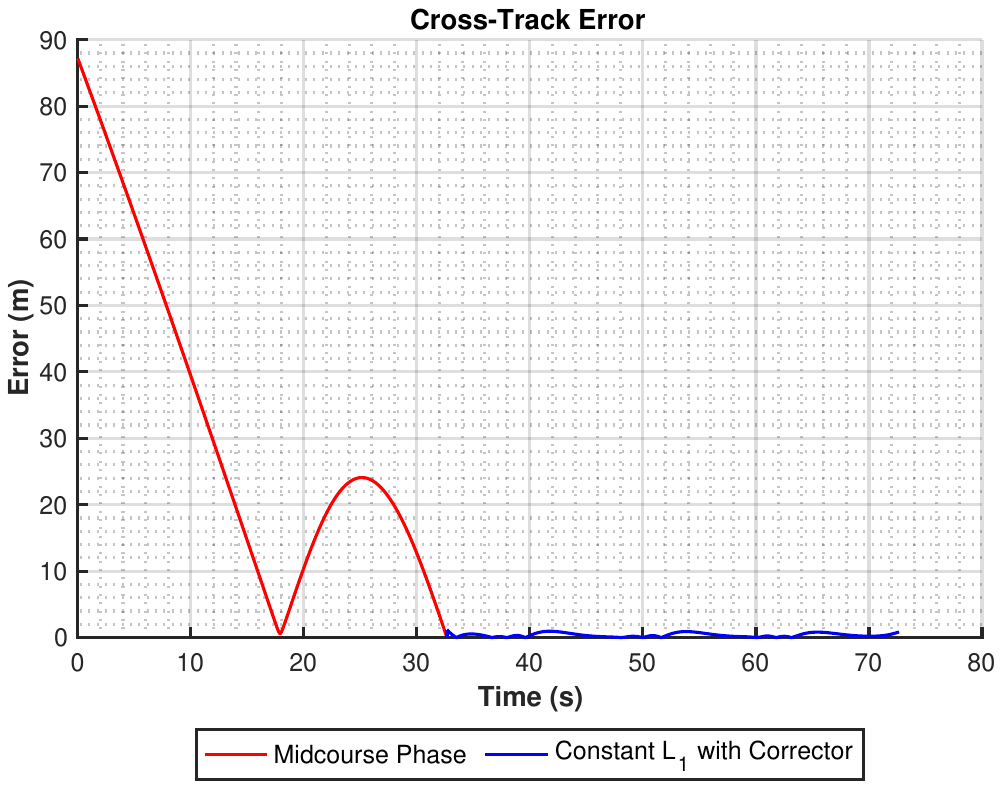}
        \caption{B. Cross-track error variation}
        \label{fig:crosswholesim}
    \end{figure}

\begin{figure}[!h]
    \centering
        \includegraphics[width=1\linewidth]{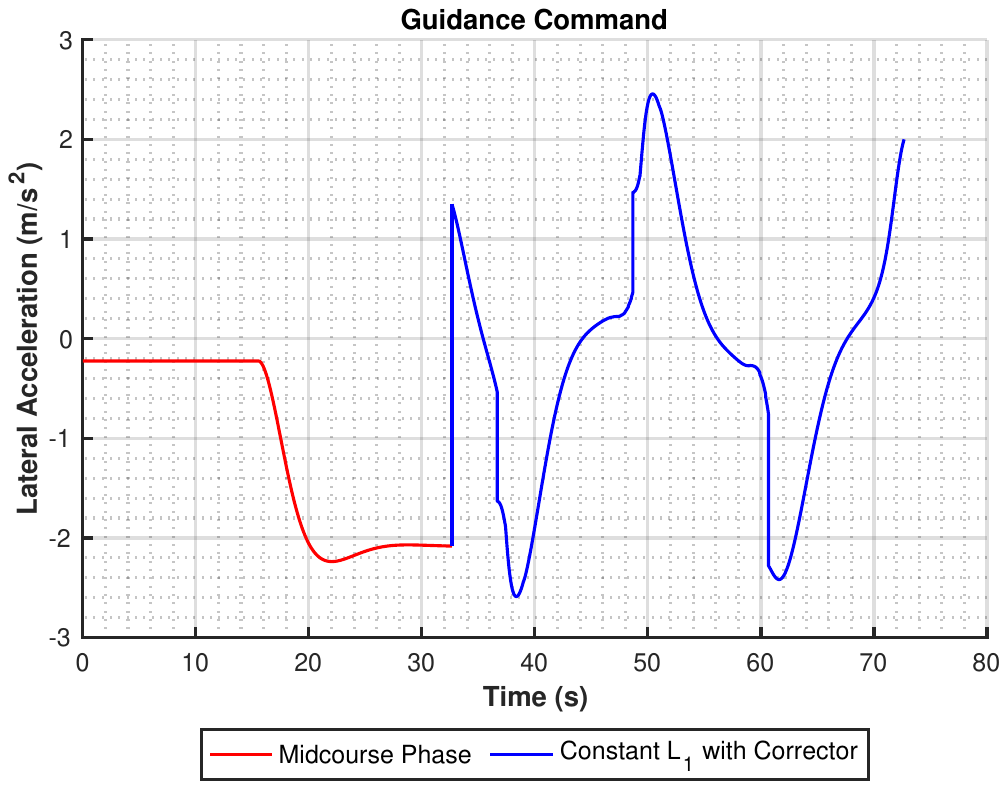}
        \caption{B. Lateral acceleration variation}
        \label{fig:accwholesim}
    \end{figure}

The vehicle finds an appropriate intermediate circle to get to and an appropriate contact point for the midcourse phase. Once on this circle as described in Section \ref{circleselection}, it switches to the close-range tracking. Fig. \ref{fig:wholesim} shows the trajectory of all the phases of guidance. The cross-track error is plotted in Fig. \ref{fig:crosswholesim}. The distance from the start point is taken as the cross-track error for the midcourse phase. It reduces until the UxV gets to the initiation circle and rotates about it for some time before passing through the trajectory start point and switching to the close tracking phase (momentarily, the error increases and then decreases to $0$). The acceleration command is plotted in Fig. \ref{fig:accwholesim}. The constant value of the acceleration command, during the midcourse phase, is as expected from the discussion subsequent to lemma \ref{lemma1}. When the UxV rotates about the initiation circle, a change is observed. Finally, for the overall path following, the acceleration is seen to be bounded.

\section{CONCLUSION}
\label{sec:Conclusion}

The notion of a look-ahead point-based guidance strategy for following a reference path is gainfully utilized in this paper to develop a two-phase guidance algorithm for a UxV to follow a reference path starting from a far-away initial location. The reaching point on an initiation circle that passes through the start point of the reference path tangentially is first optimally selected, and the look-ahead distance is dynamically obtained to steer the UxV through a circular arc of high radius, thus ensuring minimization of the lateral acceleration requirement of the UxV during the mid-course phase. In the next phase, the close-range path following problem is addressed, for which a novel notion of a corrector point, amalgamated in the constant look-ahead distance-based guidance, helps achieve a tighter reference path following. Simulation results shown for a challenging reference path having high variation in radius of curvature over a short distance indicate the effectiveness of the developed two-phase guidance strategy in terms of both lower cross-track error and lateral acceleration requirement. Its superiority over the benchmark constant look-ahead distance-based guidance in terms of these two metrics during the terminal phase is also validated by varied simulation studies. 
Extending this guidance to 3-D reference paths could form a challenging future scope of research.



\bibliographystyle{IEEEtran}
\bibliography{sample1}

\end{document}